\documentclass[10pt,twocolumn,letterpaper]{article}

\usepackage{iccv}
\usepackage{times}
\usepackage{epsfig}
\usepackage{graphicx}
\usepackage{amsmath}
\usepackage{amssymb}
\usepackage{booktabs}
\usepackage{multirow}
\usepackage{array}
\usepackage{booktabs}
\usepackage{makecell}
\usepackage{color}
\usepackage[accsupp]{axessibility}  % Improves PDF readability for those with disabilities.

\usepackage[breaklinks=true,bookmarks=false]{hyperref}

\iccvfinalcopy % *** Uncomment this line for the final submission

\def\iccvPaperID{****} % *** Enter the ICCV Paper ID here

\ificcvfinal\fi

\begin{document}

%%%%%%%%% TITLE
\title{Affine-Consistent Transformer for Multi-Class Cell Nuclei Detection}

\author{
Junjia Huang$^{1,2 \dag}$ \and Haofeng Li$^{2 \dag}$ \and Xiang Wan$^{2}$ \and Guanbin Li$^{1 *}$ \and 
$^{1}$School of Computer Science and Engineering, Research Institute of Sun Yat-sen University in Shenzhen, \\
Sun Yat-sen University, Guangzhou, China \\
$^{2}$Shenzhen Research Institute of Big Data, The Chinese University of Hong Kong, Shenzhen, China \\
{\tt\small huangjj77@mail2.sysu.edu.cn, \{lhaof,wanxiang\}@sribd.cn, liguanbin@mail.sysu.edu.cn}
%{\tt\small huangjj77@mail2.sysu.edu.cn}
}

\maketitle
\ificcvfinal\thispagestyle{empty}\fi

\newcommand\blfootnote[1]{%
  \begingroup
  \renewcommand\thefootnote{}\footnotetext{#1}%
  \addtocounter{footnote}{-1}%
  \endgroup
}

%%%%%%%%% ABSTRACT
\begin{abstract}
Multi-class cell nuclei detection is a fundamental prerequisite in the diagnosis of histopathology. It is critical to efficiently locate and identify cells with diverse morphology and distributions in digital pathological images. Most existing methods take complex intermediate representations as learning targets and rely on inflexible post-refinements while paying less attention to various cell density and fields of view. In this paper, we propose a novel Affine-Consistent Transformer (AC-Former), which directly yields a sequence of nucleus positions and is trained collaboratively through two sub-networks, a global and a local network. The local branch learns to infer distorted input images of smaller scales while the global network outputs the large-scale predictions as extra supervision signals. We further introduce an Adaptive Affine Transformer (AAT) module, which can automatically learn the key spatial transformations to warp original images for local network training. The AAT module works by learning to capture the transformed image regions that are more valuable for training the model. Experimental results demonstrate that the proposed method significantly outperforms existing state-of-the-art algorithms on various benchmarks.\blfootnote{
$^{\dag}$Junjia Huang and Haofeng Li contribute equally to this work.\\
\indent\indent$^{*}$Guanbin Li is the corresponding author.}
\end{abstract}

%%%%%%%%% BODY TEXT
\section{Introduction}
\label{sec:intro} 
A major task of pathologists is to make a diagnosis with digital pathological images, which are obtained by scanning tissue slides with a whole-slide scanner~\cite{niazi2019digital, van2021deep}. In this process, a pathologist is required to provide the grading of tumors and to classify benign and malignant diseases~\cite{nawaz2016computational, gurcan2009histopathological}, by locating and identifying certain histological structures such as lymphocytes, cancer nuclei, and glands. In some applications, instead of locating pixels on each nucleus boundary, it could be useful to quantify the different categories of cells. For example, the counts of tumor cells and lymphocytes have been utilized as an effective prognostic marker~\cite{fridman2012immune}. Thus, in this paper, we do not focus on predicting the nucleus sizes or boundaries, but only aim at inferring the types and rough locations of cell nuclei in digital slide images, following the previous work~\cite{abousamra2021multi}. 

\begin{figure}[!t]
  \centering
    \includegraphics[width=0.47\textwidth]{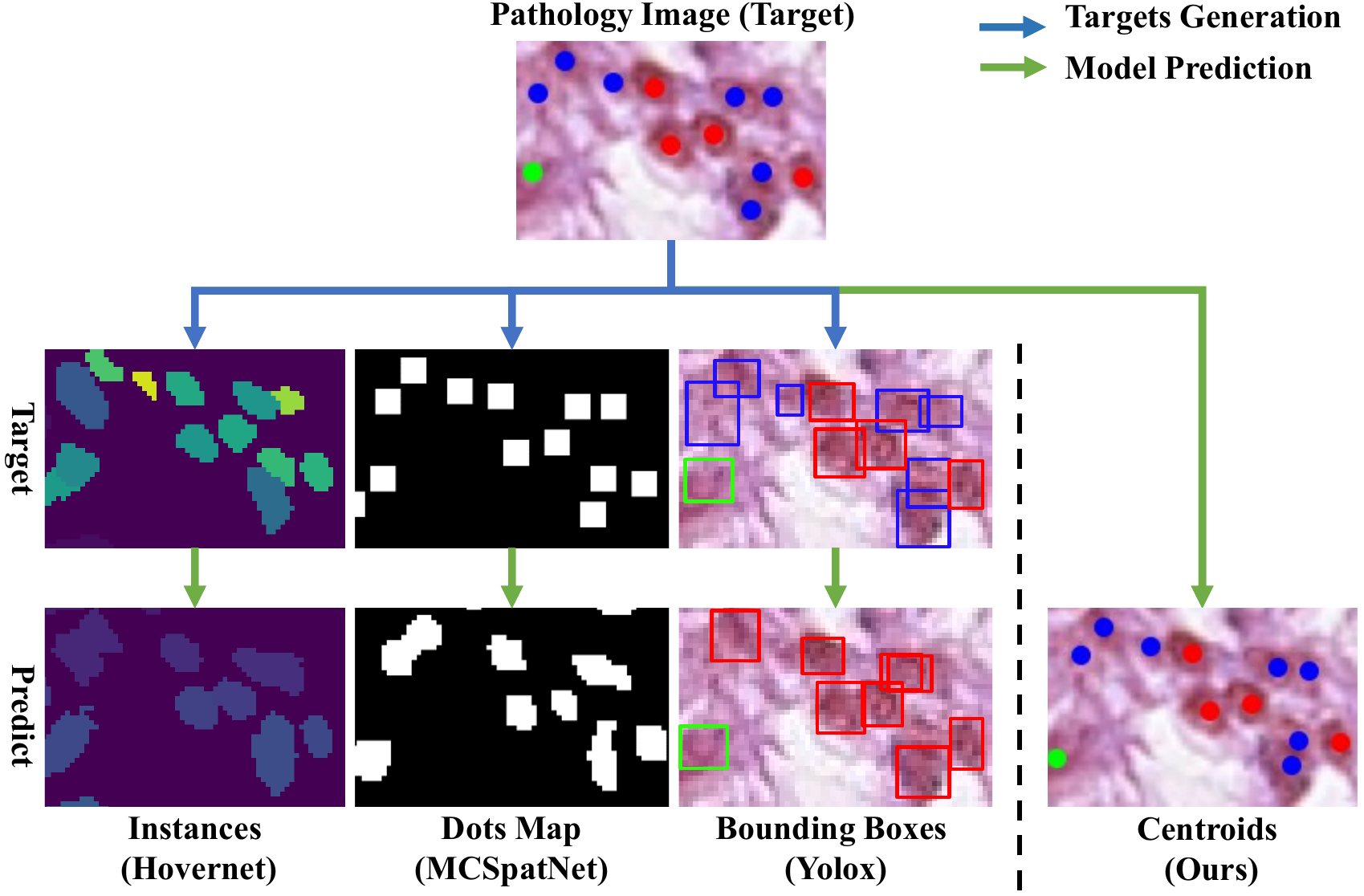}
    \caption{The visual comparison of predictions and training targets between existing methods and ours. Different types of nuclei are marked by red, green and blue boxes or centroids. Our method can predict a sequence of position coordinates and categorical labels of nuclei directly from an input pathological image.}  
  \label{intro}
\end{figure}

In the early stage, automatic nucleus detection and classification have been achieved by handcrafted features based methods\cite{plissiti2010automated,al2009improved, xu2016automatic,arteta2012learning}. These methods lack sufficient accuracy and generalization, while deep learning (DL) models can tackle these issues via learning robust representations.
 
For nuclei detection, existing DL methods can be divided into three groups, according to the different forms of prediction targets. 
As Figure~\ref{intro} shows, the first group is to outline the contour or to locate the region of each single nucleus via pixel-wise prediction~\cite{graham2019hover,doan2022sonnet,he2021cdnet,valanarasu2021medical,lou2022pixel,lou2023multi}. These methods rely on the high-quality boundary annotation of nuclei that are expensive and time-consuming. 
The second group~\cite{hofener2018deep, abousamra2021multi, sugimoto2022multi} is to pixel-wisely predict centroids or dilated centroids (`Dots Map' in Figure~\ref{intro}), by converting the detection into a binary segmentation task. Due to the diversity of cell density, the boundaries between adjacent nuclei are often confused, which makes these methods fail to segment adherent nuclei and results in missed detection.
The third group is to predict the bounding boxes of nuclei~\cite{sun2021srpn, liang2022region, zhou2021ssmd} based on the anchors of pixels, but the performance of these methods are affected by anchor parameters and post-processing. Adjacent nuclei with unclear boundaries increase the difficulty of detecting bounding boxes. Thus, we propose to convert the nuclei detection into a task of directly predicting a set of cell positions and categories. 

Besides, the diversity of image scale and nuclei density causes more difficulties in the detection and classification tasks. Higher magnification levels or scaling factors could lead to a smaller field of view and more sparse distribution of nuclei. We claim that it is essential to develop a robust model for different image scales. Some existing works~\cite{zeng2019ric, xu2015stacked} employ multi-scale deep learning architectures or simply unify the scale by dividing patches, which does not take the prediction consistency among multiple scales into consideration. 

To avoid the synthesis of indirect learning targets, we consider formulating the nucleus localization and classification problem as a sequence generation task. A transformer-based framework is adopted to decode a list of position coordinates and category labels of nuclei in a direct manner. To adapt to diverse scales, we further split the transformer framework into two network branches, a local network and a global one, which aim to infer global-scale images and their local-scale views, respectively. 
The local network is not only supervised by the ground-truth annotations but also guided by the global network that captures broader contextual information from the large-scale input. Thus, the well-trained networks could accommodate to diverse fields of view. To compute the training losses, a matching algorithm is utilized to assign each target nucleus to a nucleus proposal in the predicted sequence. Importantly, we claim that not all local image regions are equal for training a scale-consistent nuclei detection model. Therefore, we propose a novel adaptive affine transformer that predicts a series of affine transformation parameters to harvest the key local-scale inputs for improving the global-local training. Since our proposed framework is trained to deal with various fields of view and distributions of cells, it has the potential to well separate the densely distributed nuclei from each other and to reduce the missed detection rate. 

In short, our major contributions are summarized as three folds:
\begin{itemize}
	\item We introduce a novel Adaptive Affine Transformer that automatically learns to augment effective multi-scale samples for training.
    \item We propose an Affine-Consistent Transformer framework for nuclei detection. Its local branch learns to output a set of nucleus-level predictions with small-scale inputs, guided by the global branch with a large-scale input.
    \item We conduct extensive experiments and demonstrate the state-of-the-art performance of our method on three widely-used benchmarks.
\end{itemize}

\section{Related Work}
\label{sec:related work}
%\subsection{Nuclei Detection and Classification}
\noindent\textbf{Nuclei Detection and Classification}
Many methods have been developed to locate and identify cell nuclei. According to the different representations of prediction targets, they can be divided into three types: instance based, dots map based and bounding box based methods. The instance based methods~\cite{graham2019hover, doan2022sonnet, he2021cdnet,yu2023diffusion} first use neural networks to output pixel-level predictions such as semantic segmentation maps and distance maps, and then obtain the mask of each single nucleus via some post-processing methods like watershed algorithm~\cite{doan2022sonnet}. Some works \cite{zhang2018panoptic,loh2021deep,paing2022instance} detect nuclei with the generic instance segmentation methods proposed for natural images. However, these instance-based approaches require expensive pixel-level annotations of each nucleus boundary, while our method only needs lower-cost annotations of nucleus position for the detection task.

For the dots map based methods, they either regress a pixel-wise density map to locate the nuclei at the peak \cite{xie2015beyond, hofener2018deep, feng2021mutual, abousamra2021multi, wang2022global,wu20223d}, or classify image patches with sliding windows \cite{sirinukunwattana2016locality, xu2015stacked}. Abousamra \textit{et al.} \cite{abousamra2021multi} formulate the nuclei detection problem as a binary segmentation task of dilated nucleus centroids, while Wang \textit{et al.} \cite{wang2022global} extract local features and performs adversarial alignment for domain adaptive nuclei detection. Although compelling models have been proposed, these dots map based methods could fail to separate two adjacent nuclei when dealing with intensively distributed cells. Wu \textit{et al.} \cite{wu20223d} detect 3D centroids by estimating the intensity peaks of voting regions, which is different from our method that updates and outputs the centroid coordinates directly. 

\begin{figure*}[!t]
  \centering
    \includegraphics[width=0.8\textwidth]{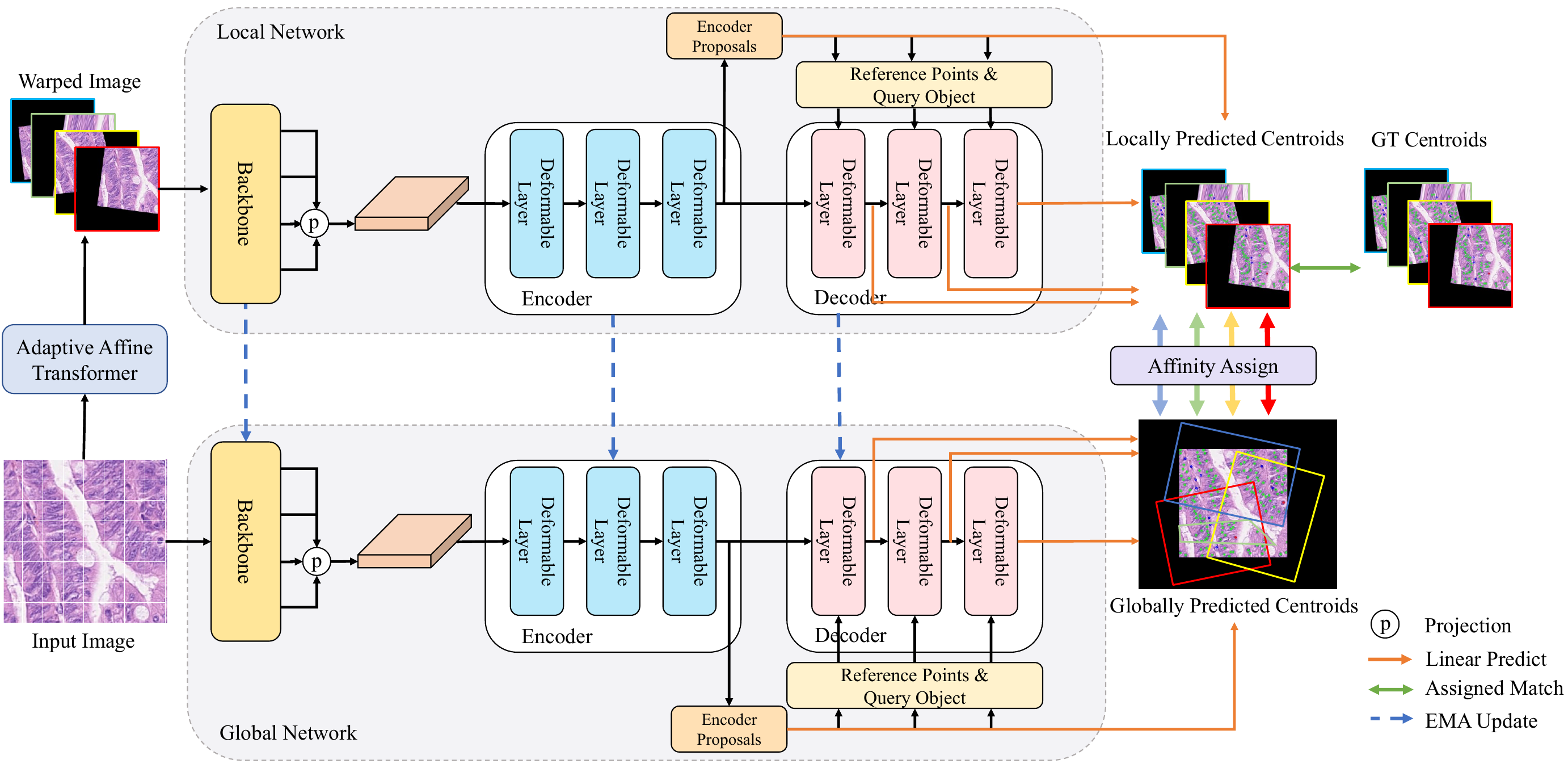}
    \caption{The overall framework of Affine-Consistent Transformer. An input training image is sent through an Adaptive Affine Transformer module to generate a series of affine transformed images. The original image and the transformed images are fed into two associated networks respectively to produce two relevant point sets with categorical scores. After one-to-one matching these two point sets, the Hungarian loss is calculated to update the local network. The output of the global network is used to coordinate the scale consistency of the local network. The global network is updated via the exponential mean average (EMA) manner.}  
  \label{framework 1}
\end{figure*}

Some other methods~\cite{yousefi2019transfer, sun2021srpn, li2021detection,wang2022deep,liang2022region, wu2021rcnn} utilize the bounding boxes of cells as training targets. Sun \textit{et al.} \cite{sun2021srpn} compute discriminative features based on similarity learning to boost the classification performance, while Liang \textit{et al.} \cite{liang2022region} propose a GA-RPN module integrating the guided anchoring (GA) into a region proposal network (RPN) to generate more suitable object proposals. Wu \textit{et al.} \cite{wu2021rcnn} detect nuclei centroids with an RCNN-SliceNet that still depends on the pipeline of producing and suppressing region proposals. These methods usually take a large number of anchor boxes as candidates and adopt the non-maximum suppression (NMS) algorithm as post-processing to screen out the highly overlapping boxes. In this paper, we avoid the tedious inference process of existing methods, and exploit transformers to directly decode the positions and category scores of nuclei. 

%\subsection{Transformer-based Object Detection}
\noindent\textbf{Transformer-based Object Detection}
Object detection aims to predict the bounding boxes and category labels of objects in an image. Transformer-based methods~\cite{carion2020end, sun2021rethinking, liu2022dabdetr, zhang2022dino, li2022view,huang2023prompt} view object detection as a set prediction problem, using transformer modules~\cite{vaswani2017attention} to directly output a final set of object-level predictions without further post-processing.  Carion~\textit{et al.} first propose a fully end-to-end object detector DETR~\cite{carion2020end} but it suffers from slow convergence and limited spatial resolution of features. In follow-up works, Zhu \textit{et al.} propose Deformable DETR~\cite{zhu2021deformable} attending to a small set of key sampling points instead of all possible pixels. Different from existing transformer-based detection models, we develop a new transformer framework that not only produces affine transformation matrices for learnable augmentation, but also adapts to nuclei detection via predicting the nucleus centroids as a sequence of points.

\section{Methodology}
\label{sec:method}

In this paper, we propose an Affine-Consistent Transformer (AC-Former). The workflow of the proposed AC-Former is shown in Figure~\ref{framework 1}. During the training, a local and a global networks cooperate with each other. The local network is trained by both the nucleus centroids from the warped images and the predicted centroids from the global network to ensure the scale consistency. The global network is continuously updated by the local network via the exponential moving average (EMA) strategy. We first introduce the proposed AC-Former framework and then describe the essential designs: an adaptive affine transformer and the local-global network architecture.

\begin{figure*}[!t]
  \centering
    \includegraphics[width=0.8\textwidth]{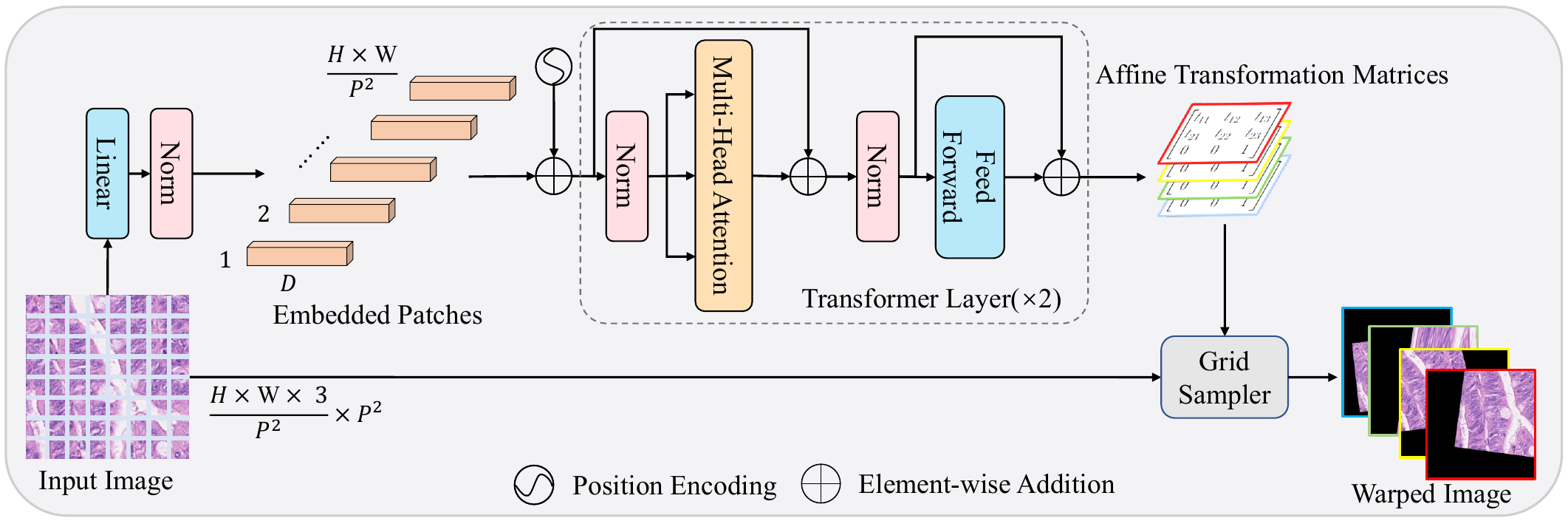}
    \caption{The detailed architecture of the Adaptive Affine Transformer. The affine transformation matrices are generated by self-attention modules, and will be adopted to perform affine warping on input images with a grid sampler. The overall structure is derivable and it will learn to automatically augment image patches that benefit the model training.}
    \label{framework 2}
\end{figure*}

\subsection{Affine-Consistent Transformer}
The proposed method learns to directly produce the centroid locations of nuclei with the corresponding confidence scores. Given a pathological image of size $H\times W\times 3$, we use the proposed adaptive affine transformer to generate warped images $I_i, i\in \{1,\cdots,M\}$. And then the local network locates the nucleus positions to generate the coordinates and category scores from the encoder and decoder. In the training stage, for the $i^{th}$ warped image, $I_i$ the local network outputs $D+1$ sets: $y^i=\{y^{ij}|j\in\{0, \cdots, D\}\}$, where $D$ is the number of layers in the decoder and $y^{ij}$ is a set of predicted centroids with coordinates and categorical scores. $y^{i0}$ is the encoder output while $y^{ij}$ with $j>0$ is the $j^{th}$ decoder output. Meanwhile, the global network infers the whole input image to obtain the globally predicted results $\Bar{y}$. Thus, the overall loss is defined as:
\begin{equation}
  \mathcal{L}(y,\hat{y},\Bar{y}) = \frac{1}{M}\sum_{i=1}^M \sum_{j=0}^D (\mathcal{L}_{m}(y^{ij}, \hat{y}^{i}) + \alpha \mathcal{L}_{m}(y^{ij}, \Bar{y}^{ij})),
  \label{eq:1}
\end{equation}
where $\mathcal{L}_m$ denotes the Hungarian loss~\cite{carion2020end} between one-to-one aligned nuclei, $\hat{y}$ is the ground truth of warped centroids. $\hat{y}^i$ represents the result of aligning the ground truth with the $i^{th}$ locally warped image, while $\Bar{y}^{ij}$ is obtained by aligning the $j^{th}$ globally predicted set with the $i^{th}$ distorted image. $\alpha$ is a weight term to balance the global branch loss and the local loss based on ground-truths. 

We introduce the prediction of the global network for supervision to enhance the spatial scale consistency of the trained model. In the centroids sets predicted by the global network, since the redundancy can not be eliminated by the non-maximum suppression (NMS) \cite{neubeck2006efficient}, we use the maximum category probability as the evaluation scores of centroids to select candidates with a threshold $t$. Only the centroids whose maximum category probability is not smaller than $t$ are reserved. Empirically we set $t=0.3$. During the inference stage, we simply employ the global network to infer a testing image, and produce $D+1$ sets of nucleus centroids. Only the last set is adopted as the final prediction of the overall method.

\subsection{Adaptive Affine Transformer}
Since the distribution of nuclei is not uniform, local views obtained by random transformations are not equally beneficial for the training. Thus, we propose to learn to synthesize the local views with a trainable model. To embrace long-range contexts, we develop a transformer-based model, Adaptive Affine Transformer (AAT) to analyze the input image and automatically generate the parameters of an affine transformation.

As shown in Figure~\ref{framework 2}, the proposed adaptive affine transformer divides an input image into $\frac{HW}{P^2}$ patches, feeds the flattened patches to a linear projection layer, and obtains the embedded patches of size $\frac{HW}{P^2}\times E$, where $E$ denotes the number of embedding dimensions and $P$ is the size of each image patch. After that, the embedded patches are added with their sinusoidal position embeddings, and then passed through a transformer \cite{vaswani2017attention} encoder with 2 layers. These transformer layers compute the global dependency between image patches with the multi-head attention mechanism. The output is linearly projected into $M$ affine transformation matrices $\{A_1, ... , A_M\}$. We do not adopt perspective transformations but only affine transformations. The pointwise transformation process can be formulated as :
\begin{equation}
  \sigma(u,v,A) = A \begin{bmatrix}
                 u \\ v \\ 1
                \end{bmatrix}
                = \begin{bmatrix} 
                t_{11} & t_{12} & t_{13} \\
                t_{21} & t_{22} & t_{23} \\
                0   & 0   & 1   \\
                \end{bmatrix}
                \begin{bmatrix}
                u \\ v \\ 1
                \end{bmatrix}
                = \begin{bmatrix}
                u'\\v'\\ 1
                \end{bmatrix},
  \label{eq:2}
\end{equation}
where $u$ and $v$ are the coordinates of a sampled point in the input image. $u'$ and $v'$ are the target coordinates after the affine transformation with $A$. 

To avoid generating unidentifiable images of low-quality, we constrain the matrix parameters in a moderate range. For scaling factors in the matrix, $t_{11}, t_{22}$, we clamp them into [0.2, 2]; for translation, $t_{13}, t_{23}$ are clamped into [-0.5, 0.5]; for rotation \& shear, $t_{12}, t_{21}$ are clamped into [-1, 1]. Then we feed these $M$ matrices with an input image into a differentiable grid sampler \cite{jaderberg2015spatial} to yield $M$ distorted images. Let $V(i,j)$ return the pixel value at the position $(i,j)$ of the input image (in Figure~\ref{framework 2}). $I(u',v')$ denotes the pixel value at the position $(u',v')$ of the warped image, and can be computed by bilinear interpolation:
\setlength{\arraycolsep}{2.5pt}
\begin{equation}
    I(u',v') = \begin{bmatrix}
                    1-u & u
               \end{bmatrix}
               \begin{bmatrix}
                  V(\lfloor u \rfloor, \lfloor v \rfloor) &  V(\lfloor u \rfloor, \lceil v \rceil) \\
                  V(\lceil u \rceil, \lfloor v \rfloor) &  V(\lceil u \rceil, \lceil v \rceil)
               \end{bmatrix}
               \begin{bmatrix}
                  1-v \\ v
               \end{bmatrix}
\label{eq:3}
\end{equation}

To produce the pixel values of $M$ warped images, each integer coordinate of the $i^{th}$ warped image is set to $(u',v')$ and its corresponding $(u,v)$ can be solved via Eq.~\ref{eq:2}, given the $i^{th}$ affine transformation matrix $A_i$. Then the pixel value at the position $(u',v')$ of a warped image can be calculated via Eq.~\ref{eq:3}. To yield $\hat{y}^i$ in Eq.~\ref{eq:1}, that is to say, to align $\hat{y}$ with $i^{th}$ distorted image, we set each centroid coordinate in $\hat{y}$ to $(u,v)$, and compute the corresponding $(u',v')$ via Eq.~\ref{eq:2} with $A_i$. The resulted coordinates $(u',v')$ out of the range $([0,H],[0,W])$ are removed from $\hat{y}^i$. Computing $\Bar{y}^{ij}$, namely, aligning the $j^{th}$ predicted set of the global network with the $i^{th}$ warped image, is performed in a similar manner. The process can be formulated as: $\Bar{y}^{ij} = \{\sigma(\Bar{y}^{j}_q, A_i) | q\in\{1, \cdots, |\Bar{y}^{j}|\} \}$ where $\Bar{y}^{j}_q$ denotes the $q^{th}$ centroid coordinate of the globally predicted set $\Bar{y}^{j}$. The proposed adaptive affine transformer module enables the network to adaptively learn enhanced features that are robust for spatial transformations. 

\subsection{Global-Local Network Architecture}
\noindent\textbf{Backbone} As illustrated in Figure~\ref{framework 1}, the global or the local network consists of 3 components: a backbone, an encoder and a decoder. We adopt ConvNeXt-B~\cite{liu2022convnet} as the backbone that acts as a deep feature extractor. The backbone yields a list of feature maps of different scales.

\noindent\textbf{Encoder} 
The encoder has 3 deformable attention~\cite{zhu2021deformable} layers and 2 fully-connected (FC) layers. The input of the encoder is a stack of multi-scale feature maps output by the backbone. For each attention layer in the encoder, its input and output have the same shape, and its query elements are set to all pixel-level feature vectors in its multi-scale input feature maps. After the attention layers, each feature vector is separately sent into the FC layers, which produce the categorical scores and the coordinate offsets relative to the feature position. Only $n$ feature vectors with the highest confidence are preserved as object query embeddings and their coordinates are recorded as the reference points.

\noindent\textbf{Decoder} The decoder has 3 layers and each layer contains a deformable attention module and 2 FC layers. Different from the encoder, the deformable attention module in the decoder takes the object queries from the encoder as query elements. After the attention enhances the object query embeddings, the two FC layers predict a 2D offset and categorical scores for each object query, respectively. The 2D offset is added to and updates the corresponding reference point. For deep supervision~\cite{shen2017dsod}, the 3 decoder layers provide 3 sets of predictions, respectively.

The loss is calculated between each predicted set of centroid proposals and a set of target centroids that could be ground truths or predicted by the global network. The number of centroid proposals $C$ is far more than that of target centroids $T$. Let $y_c=\{(u_c,v_c),c_c\}$ denote a centroid proposal, and $y_t=\{(u_t,v_t),c_t\}$ denote a centroid target. A centroid proposal or centroid target consists of the coordinates and category scores. A pair-wise cost matrix $\mathcal{E}$ is computed by measuring the cost between each proposal and each target:
\begin{equation}
    \mathcal{E}(y_c, y_t) = \beta_1||(u_c,v_c)-(u_t,v_t)||_2^2+\beta_2 l_{focal}(c_c, c_t),
\end{equation}
where $l_{focal}(\cdot)$ denotes the Focal Loss \cite{lin2017focal}. $\beta_1$ and $\beta_2$ provide a balance between the position regression and classification. The Focal Loss is used to mitigate the class imbalance of the nuclei classification task. It is defined as:
\begin{equation}
    l_{focal}(c_c, c_t) = \frac{1}{K}\sum_{k=1}^K -\lambda_1 (1-c_{ck}\cdot c_{tk})^{\lambda_2}c_{tk}log(c_{ck}),
\end{equation}
where $\lambda_1$ and $\lambda_2$ are the balanced factors and the focusing parameter, K denotes the number of classes. Then we conduct the association with the Hungarian algorithm \cite{kuhn1955hungarian} based on the $C\times T$ cost matrix $\mathcal{E}$, and obtain $T$ matching positives and $C-T$ negatives. Our goal is to narrow the coordinate and categorical difference between the positive proposals and their corresponding target, and to amplify the categorical difference between the negatives and the positives. For all positive and negative samples in a warped image $I_i$, the loss is formulated as: 
\begin{align}
  &\mathcal{L}_{m}(y^{ij}, \hat{y}^i) = \frac{1}{T}(\omega_1\sum^T_{p=1}||(u^{ij}_p,v^{ij}_p)-(\hat{u}^i_p,\hat{v}^i_p)||_2^2  \nonumber \\
        &+ \omega_2l_{focal}(c^{ij}_{p}, \hat{c}^i_{p}) 
        +\omega_3\sum^C_{n=T+1}l_{focal}(c^{ij}_{n},\hat{c}^i_{n})),
\label{eq:7}
\end{align}
where $\omega_1, \omega_2$ and $\omega_3$ are weight terms. $\{(u^{ij}_p,v^{ij}_p),c^{ij}_p\}$ denotes the $p^{th}$ matching positive centroid of the $j^{th}$ predicted set in the $i^{th}$ warped image and $\{(\hat{u}^i_p, \hat{v}^i_p), \hat{c}^i_p\}$ is the corresponding target in the ground truth $\hat{y}$ after the $i^{th}$ affine transformation. For the negative proposals, we define their classification target $\hat{c}_n^i$ as a new empty category. The loss between the local and the global predictions can be calculated as $\mathcal{L}_{m}(y^{ij},\Bar{y}^{ij})$ in Eq.~\ref{eq:3}, in a way similar to Eq.~\ref{eq:7}. 

In the training stage, the encoder and each of the 3 decoder layers predict a set of centroids separately, resulting in 4 (1+3) sets of centroids. which are all used to compute the loss. During the inference, we take the output of the last decoder layer as the final prediction. The hyper-parameters $\alpha, \beta 1, \beta 2, \lambda 1, \lambda 2, \omega 1, \omega 2, \omega 3$ are set based on MMDetection~\cite{mmdetection} and Deformable DETR~\cite{zhu2021deformable}, and do not need any complicated tuning.

\renewcommand{\arraystretch}{1.12}
\begin{table*}[!t]
    \centering
    \caption{Results on three benchmarks, CoNSep, BRCA-M2C and Lizard. For each dataset, we report the F-score of each class ($F_c^k$), the mean F-score over all classes ($\overline{F_c}$) and the detection F-score ($F_d$). $F_c^{Infl.}$, $F_c^{Epi.}$, $F_c^{Stro.}$, $F_c^{Neu.}$, $F_c^{Lym.}$, $F_c^{Pla.}$, $F_c^{Eos.}$ and $F_c^{Con.}$ denote the F-socre for the inflammatory, epithelial, stromal, neutrophils, lymphocytes, plasma, Eosinophil and connective tissue cells, respectively. For each row, the best method is in bold type and the second best method is underlined.}
    \label{tab:sota}
    \setlength{\tabcolsep}{0.8mm}{
    \small 
    %\footnotesize
    \begin{tabular}{c|l|cccccccc}
    \Xhline{1pt}
         ~ & \multirow{2}*{F-score$\uparrow$} & \multirow{2}*{Hovernet \cite{graham2019hover}} & \multirow{2}*{DDOD \cite{chen2021disentangle}} & \multirow{2}*{TOOD \cite{feng2021tood}} &  \multirow{2}*{MCSpatNet \cite{abousamra2021multi}} & \multirow{2}*{SONNET \cite{doan2022sonnet}} & \multirow{2}*{DAB-DETR \cite{liu2022dabdetr}} &  \multirow{2}*{\makecell[c]{UperNet-\\ConvNeXt\cite{liu2022convnet}}} & \multirow{2}*{\makecell[c]{AC-Former\\(Ours)}} \\
         ~ & ~ & ~ & ~ & ~ & ~ & ~ &  ~ & ~ & ~ \\
        \hline
        ~ & ~ & 2019 & 2021 & 2021 & 2021 & 2022 & 2022 & 2022 & - \\
         \hline
        \multirow{5}*{\rotatebox{90}{CoNSeP}} 
          & \ $F_c^{Infl.}$ & 0.514 & 0.516 & \underline{0.622} & 0.583 & 0.563 & 0.531 & 0.618 & \textbf{0.635} \\ 
        ~ & \ $F_c^{Epi.}$ & 0.604 & 0.436 & 0.616 & 0.608 & 0.502 & 0.440 & \underline{0.625} & \textbf{0.635} \\ 
        ~ & \ $F_c^{Stro.}$ & 0.391 & 0.429 & 0.382 & 0.527 & 0.366 & 0.443 & \underline{0.542} & \textbf{0.568} \\
        ~ & \ $\overline{F_c}$ & 0.503 & 0.494 & 0.540 & 0.573 & 0.477 & 0.471 & \underline{0.595} & \textbf{0.613} \\ 
        ~ & \ $F_d$ & 0.621 & 0.554 & 0.608 & \underline{0.722} & 0.590 & 0.619 & 0.715 & \textbf{0.739} \\ 
        \hline
        \multirow{5}*{\rotatebox{90}{BRCA-M2C}} 
          & \ $F_c^{Infl.}$ & \underline{0.454} & 0.394 & 0.400 & 0.424 & 0.343 & 0.437 & 0.423 & \textbf{0.474} \\ 
        ~ & \ $F_c^{Epi.}$ & 0.577 & 0.544 & 0.559 & 0.627 & 0.411 & 0.634 & \underline{0.636} & \textbf{0.637} \\ 
        ~ & \ $F_c^{Stro.}$ & 0.339 & 0.373 & 0.315 & \textbf{0.387} & 0.281 & \underline{0.380} & 0.353 & 0.344 \\ 
        ~ & \ $\overline{F_c}$ & 0.457 & 0.437 & 0.425 & 0.479 & 0.345 & \underline{0.484} & 0.471 & \textbf{0.485} \\ 
        ~ & \ $F_d$ & 0.74 & 0.659 & 0.662 & \underline{0.794} & 0.653 & 0.705 & 0.785 & \textbf{0.796} \\ 
        \hline
        \multirow{7}*{\rotatebox{90}{Lizard}} 
          & \ $F_c^{Neu.}$ & \underline{0.210} & 0.025 & 0.029 & 0.105 & 0.09  & 0.142 & 0.205 & \textbf{0.270} \\ 
        ~ & \ $F_c^{Epi.}$ & 0.665 & 0.584 & 0.615 & 0.601 & 0.599 & 0.653 & \underline{0.714} & \textbf{0.788} \\ 
        ~ & \ $F_c^{Lym.}$ & 0.472 & 0.342 & 0.404 & 0.457 & 0.538 & 0.544 & \underline{0.611} & \textbf{0.690} \\ 
        ~ & \ $F_c^{Pla.}$ & \underline{0.376} & 0.130 & 0.152 & 0.228 & 0.370 & 0.356 & 0.333 & \textbf{0.475} \\ 
        ~ & \ $F_c^{Eos.}$ & 0.367 & 0.124 & 0.157 & 0.220 & 0.365 & 0.295 & \underline{0.403} & \textbf{0.450} \\ 
        ~ & \ $F_c^{Con.}$ & 0.492 & 0.347 & 0.383 & 0.484 & 0.143 & 0.559 & \underline{0.578} & \textbf{0.671} \\ 
        ~ & \ $\overline{F_c}$ & 0.430 & 0.259 & 0.290 & 0.349 & 0.351 & 0.425 & \underline{0.474} & \textbf{0.557} \\ 
        ~ & \ $F_d$ & 0.729 & 0.561 & 0.606 & 0.713 &  0.682 & 0.656 & \underline{0.764} & \textbf{0.782} \\
    \Xhline{1pt}
    \end{tabular}
    }
\end{table*}

\section{Experiments}
\subsection{Experimental Settings}
\noindent\textbf{Datasets.} We evaluate the proposed approach on three publicly available datasets,  {CoNSeP} \cite{graham2019hover},  {BRCA-M2C} \cite{abousamra2021multi} and  {Lizard} \cite{graham2021lizard}. 
 {CoNSeP} is a colorectal nuclear dataset, consisting of 41 H\&E stained image tiles from 16 colorectal adenocarcinoma whole-slide images (WSIs). 
 {BRCA-M2C} is a breast cancer dataset and consists of 120 image tiles from 113 patients. Both  {CoNSeP} and  {BRCA-M2C} contain three types of cells:  inflammatory, epithelial, or stromal.
 {Lizard} has 291 histology images of colon tissue from six different dataset sources, containing nearly half a million labeled nuclei in H\&E stained colon tissue.  {Lizard} provides nucleus-level class labels for epithelial cells, connective tissue cells, lymphocytes, plasma cells and neutrophils.
 {CoNSeP} and  {Lizard} contain the annotated contour masks of nuclei while  {BRCA-M2C} only has the labels of centroids. To run instance based and bounding box based methods on  {BRCA-M2C}, we follow the work~\cite{abousamra2021multi} to apply the SLIC \cite{achanta2012slic} algorithm to generate superpixels as instances. We split the fully labeled samples into training, validation, and test sets, following the official partition \cite{graham2019hover, abousamra2021multi, graham2021lizard}.

\noindent\textbf{Evaluation metrics. }
We follow the work~\cite{graham2019hover} and use F-score to evaluate the detection and classification tasks. A higher F-score means better performance. For the detection, we compute the Euclidean distance between each predicted centroid and GT to yield a cost matrix. Then we run the Hungarian algorithm~\cite{kuhn1955hungarian} with the cost matrix to obtain the paired results, and set the pairs beyond 6 pixels to unpaired samples. The predicted centroids belonging to some pair are correctly detected centroids ($TP_d$, $d$ for detection) while the rests are overdetected predicted centroids ($FP_d$). The GT centroids without matched predictions are called misdetected GT ($FN_d$). The detection F-score is calculated with the size of the above sets of nuclei: $F_d=\frac{2TP_d}{2TP_d+FP_d+FN_d}$.

For the classification task with $K$ classes, $TP_d$ are further split into the following subsets: correctly classified centroids of Type $k$ $(TP^k_c)$, incorrectly classified centroids of Type $k$ ($FP^k_c$) and incorrectly classified centroids of types other than Type $k$ ($FN^k_c$). The classification F-score is defined as: $F^k_c = \frac{2TP^k_c}{2(TP^k_c+FP^k_c+FN^k_c)+FP_d+FN_d}$.
        
\noindent\textbf{Implementation details.}
Our implementation is based on MMDetection~\cite{mmdetection}. We use AdamW \cite{loshchilov2018decoupled} optimizer with a learning rate of $2^{-4}$ to train models. We load the ImageNet \cite{deng2009imagenet} pre-trained weights for initializing the ConvNeXt-B backbone and the embedding dimension $E$ is set to 128. For training and inference, we remove the centroid proposal whose maximum category score is smaller than a threshold of 0.5 and no more than $n$ proposals are preserved. $n$ is usually set to 1000. During the training, the network is trained with only the local branch loss in early steps and then the overall loss function (Eq.~\ref{eq:1}) is optimized with $\alpha=0.1$. During the inference, we adopt the global network for prediction with sliding windows. We apply the multi-scale training with sizes between 600 and 800, and infer the image with a size of $800\!\times\!800$. More details are listed in the supplementary material. 

\renewcommand{\arraystretch}{1.12}
\begin{table*}
  \centering
    \caption{The results of ablation study. The results are obtained on  the  {CoNSeP} and  {Lizard} datasets. `BL' means training the single-branch baseline in our method, with original pathological images. `RC', `RR' and `RT' denote the random crop, random rotation and random affine transformation strategies for synthesizing local views. `AAT' is the Adaptive Affine Transformer and `GL' is the supervised loss based on the global branch.}
    \setlength{\tabcolsep}{1.8mm}{
  %\small 
  \footnotesize
  \begin{tabular}{c|ccccc|cccccccc}
    \Xhline{1pt}
    \specialrule{0em}{2pt}{1pt}
    \multirow{2}*{Methods} & \multicolumn{4}{c}{\quad\quad CoNSeP}  & ~ & \multicolumn{8}{c}{Lizard} \\ 
    \Xcline{2-14}{0.4pt}
    \specialrule{0em}{2pt}{1pt}
    ~ & $F_c^{Infl.}$ & $F_c^{Epi.}$ & $F_c^{Stro.}$ & $\overline{F_c}$ & $F_d$ & $F_c^{Neu.}$ & $F_c^{Epi.}$ & $F_c^{Lym.}$ & $F_c^{Pla.}$ & $F_c^{Eos.}$ & $F_c^{Con.}$ &$\Bar{F_c}$ & $F_d$ \\
    \midrule
    \midrule
    BL & 0.571 & 0.627 & 0.538 & 0.579 & 0.696 & 0.042 & 0.740 & 0.629 & 0.395 & 0.348 & 0.526 & 0.447 & 0.715 \\
    BL+RC & 0.604 & 0.625 & 0.560 & 0.596 & 0.713 & 0.011 & 0.707 & 0.625 & 0.365 & 0.335 & 0.629 & 0.445 & 0.725\\
    BL+RR & 0.595 & 0.619 & 0.545 & 0.584 & 0.704 & 0.010 & 0.743 & 0.612 & 0.389 & 0.347 & 0.633 & 0.456 & 0.732 \\
    BL+RT & 0.617 & 0.565 & 0.541 & 0.574 & 0.711 & 0.070 & 0.741 & 0.62 & 0.414 & 0.393 & 0.660 & 0.474 & 0.749\\
    BL+AAT & 0.603 & 0.627 & 0.553 & 0.594 & 0.721 & 0.234 & 0.774 & 0.659 & 0.440 & 0.428 & 0.615 & 0.525 & 0.752 \\
    \hline
    BL+RC+GL & 0.606 & \textbf{0.648} & 0.551 & 0.602 & 0.730 & 0.187 & 0.758 & 0.677 & 0.439 & 0.430 & 0.662 & 0.526 & 0.769 \\
    BL+RR+GL & 0.626 & 0.638 & 0.555 & 0.606 & 0.726 & 0.140 & 0.758 & 0.643 & 0.411 & 0.404 & 0.657 & 0.502 & 0.747 \\
    BL+RT+GL & 0.\textbf{642} & 0.562 & 0.543 & 0.582 & 0.725 & 0.174 & 0.768 & 0.664 & 0.453 & \textbf{0.454} & \textbf{0.679} & 0.532 & 0.775\\
    \makecell[c]{BL+AAT+GL \\ (Ours)} & 0.635 & 0.635 & \textbf{0.568} & \textbf{0.613} & \textbf{0.739} & \textbf{0.270} & \textbf{0.788} & \textbf{0.690} & \textbf{0.475} & 0.450 & 0.671 & \textbf{0.557} & \textbf{0.782}\\
    \specialrule{0em}{1pt}{2pt}
    \Xhline{1pt}
  \end{tabular}
  }
  \label{tab:ab}
\end{table*}

\subsection{Comparison with State-of-the-arts}
As shown in Table~\ref{tab:sota}, we compare our proposed method with the state-of-the-art approaches which can jointly segment/detect and classify cell nuclei. These approaches include the instanced based methods (Hovernet~\cite{graham2019hover}, SONNET~\cite{doan2022sonnet}), the bounding box based methods (DAB-DETR~\cite{liu2022dabdetr}, TOOD~\cite{feng2021tood}, DDOD~\cite{chen2021disentangle}), and the dots map based methods (UperNet with ConvNeXt~\cite{liu2022convnet} backbone, MCSpatNet~\cite{abousamra2021multi}). In Table~\ref{tab:sota}, the proposed method achieves the highest mean F-score in both detection and classification tasks on the benchmarks  {CoNSeP},  {BRCA-M2C} and  {Lizard}.

\begin{figure*}[!t]
  \centering
  \includegraphics[width=0.8\textwidth]{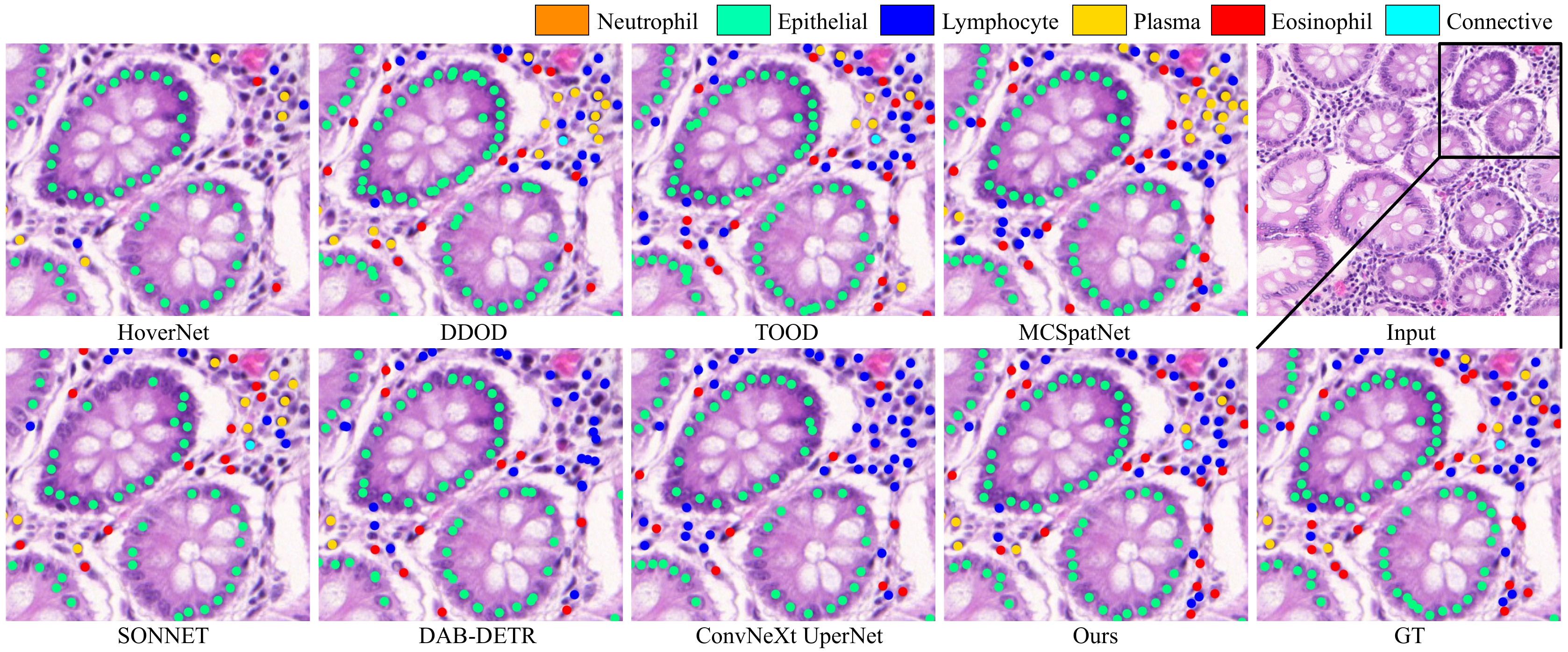}
   \caption{ Qualitative comparison on the  {Lizard} dataset. The four leftmost columns visualize the predictions of existing methods and ours. The rightmost column displays the input image and the ground truth annotations. Five types of cell are marked with dilated nucleus centroids in five different colors. As the results show, the category distribution of our method is the closest to that of the ground truths.}
   \label{fig:visual}
\end{figure*}

For the  {Lizard} dataset, our proposed method demonstrates 1.8\% F-socre in detection and 8.3\% F-score in classification higher than the second best model MCSpatNet, respectively. Some existing models show inferior results. It may be that Lizard is a newly released and challenging benchmark that has the class imbalance problem. Figure~\ref{fig:visual} presents a qualitative comparison between the state-of-the-art algorithms and the proposed network. More results are presented in the supplemental materials.

\noindent\textbf{Comparison using the Same Backbone.} Consider that a large-size of high-capacity backbone may improve or degrade the performance due to over-fitting. To fairly reproduce the existing methods in Table \ref{tab:sota}, their backbones are set following their original paper. All the backbones are pre-trained on the ImageNet. Note that in Table~\ref{tab:sota} even though DAB-DETR and UperNet use the same backbone ConvNeXt as our method, the proposed model significantly exceeds them by 2.4\%-12\% in $F_d$ on the CoNSep dataset. 

\noindent\textbf{Comparison with Bounding Box based Methods.}
Bounding box methods provide more information like cell sizes, but their labels are more expensive than the centroid labels used by our method. We only aim at locating more cells with correct labels and reducing the missing rate, which can be used to compute the counts of cells as prognostic markers~\cite{fridman2012immune}. In Table~\ref{tab:sota}, two competitive bounding-box models DDOD and DAB-DETR are compared with ours. The proposed method surpasses the two models by more than 9\% F-score in detection on the BRCA-M2C dataset. Such a performance gap unveils those centroid-based methods are superior to bounding-box based ones for nuclei detection. 

\begin{figure}[!t]
  \centering
    \includegraphics[width=0.35\textwidth]{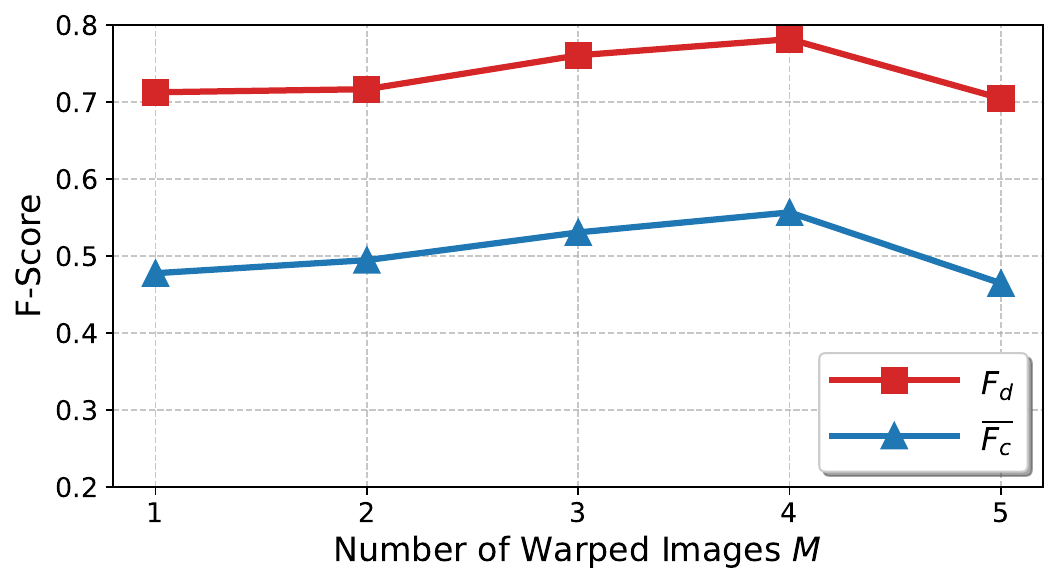}
    \caption{The effects of the number of warped images generated by the Adaptive Affine Transformer on  {Lizard} dataset. The F-scores of each nucleus category are in the supplemental materials.}
  \label{ab}
\end{figure}

\subsection{Ablation Study}

\noindent \textbf{Effectiveness of the proposed AAT module.} In Table~\ref{tab:ab}, `BL' denotes the baseline that is the global branch in our method (Figure~\ref{framework 1}). `BL+AAT' is a dual-branch model using the AAT to warp images and the EMA strategy to update the global branch. Comparing `BL+AAT' with `BL' shows that the AAT module improves the baseline by 7.8\% in $\Bar{F_c}$ and 3.7\% in $F_d$ on the Lizard dataset, which is significant. 

\noindent \textbf{Comparison with Non-learnable Data Augmentation.}
Since the AAT can learn to transform input image patches for training, we compare the AAT with other traditional non-learnable data augmentations. As Table~\ref{tab:ab} shows, we implement 3 kinds of non-learnable augmentation: Random Cropping, Random Rotation, and Random Affine Transformation, which are denoted as `+RC', `+RR', and `+RT'. 

`BL+AAT+GL' denotes our overall proposed method, while `GL' means the Global Loss (the supervision from the global branch, see the lower half of Table~\ref{tab:ab}). By replacing AAT with RC/RR/RT, we obtain the results of `BL+RC+GL', `BL+RR+GL', `BL+RT+GL'. Our proposed method outperforms the three models by 2.5\%-5.5\%  F-score in classification and 0.7\%-3.5\% F-score in detection, on the Lizard dataset. The results suggest that the proposed AAT is superior to the 3 kinds of data augmentation strategies. Interestingly, the AAT brings more significant improvements in the sub-task of cell classification, which indicates that the proposed module does synthesize useful input samples for learning more robust semantic features. 

\noindent\textbf{Effectiveness of the Global-Local Architecture.} We investigate the strengths of the global-local framework. In Table~\ref{tab:ab}, we analyze the results of different image transformation strategies, and find that the use of Global Loss $\mathcal{L}(y,\Bar{y})$ from the global network can achieve stable improvements. Specifically, on {Lizard}, `BL+AAT+GL (Ours)' surpasses `{BL+AAT}' by 3\% and 3.2\% F-scores in detection and classification, respectively. The statistics suggest that using sub-network learning from large-scale inputs can help train another sub-network to adapt to various fields of view. 

\noindent \textbf{Efficiency Analysis.}
Our AAT and dual-branch design are used only in the training stage. Thus, our method enjoys the same inference efficiency as the single-branch baseline. As Table~\ref{tab:efficiency} shows, our method avoids post-processing and takes less than 0.1 s for inferring an image, while the post-processing based methods are 3 times slower ($>$ 0.33 s). 

\noindent \textbf{Amount of Warped Images.} Figure~\ref{ab} shows the results of investigations about how the number of warped images $M$ for training a local network affects the testing F-score. The results show that setting $M$ to 4 performs the best. Setting $M$ from 2 to 4 can achieve consistent improvements in comparison to the model with $M=1$. A large number of distorted images (\textit{e.g.}, $M=5$) would degrade the results.

\renewcommand{\arraystretch}{1.0}
\begin{table}[h]
\centering
%\small
\footnotesize
\caption{Efficiency Analysis of the state-of-the-art methods and ours with a Tesla v100 (32GB), Intel Xeon Gold 6248 on CoNSeP.}
\setlength{\tabcolsep}{2mm}{ 
\begin{tabular}{c|ccc}
\Xhline{0.5pt}
~ & Time (s) & \multicolumn{2}{c} {Memory (GB)} \\
Methods & Inference+Post-process & Inference & Training \\
\Xhline{0.5pt}
Baseline (BL) & 0.097 & 11.191 & 12.298  \\ 
HoverNet & 0.021+0.376 & 11.380 & 12.892 \\
MCSpatNet & 0.058+0.287 & 2.679 & 6.228 \\
SONNET & 0.081+0.250 & 21.031 & 30.493 \\
Ours & 0.097  & 11.191 & 24.206 \\
\Xhline{0.5pt} 
\end{tabular}
}
\label{tab:efficiency}
\end{table}

\section{Conclusion}
In the paper, we propose a novel affine-consistent transformer framework that directly predicts a list of locations and categories for multi-class nuclei detection without complex post-refinements. We first introduce an Adaptive Affine Transformer module, which can automatically learn argumentation strategies to warp training input images, and enhance the model adaptability and accuracy. Next, we propose two associated networks that adapt to local-scale image views under the guidance of global-scale predictions, to boost the consistency and robustness of the model. Extensive experiments on three benchmarks have demonstrated the strengths of our overall framework AC-Former and the proposed AAT module on nuclei detection and classification tasks. 

The limitation is that our model could fail to locate incomplete nuclei that are split at image boundaries due to the lack of contextual information. To solve the issue, we may try to crop highly-overlapping image patches and stitch their results better in future work.

\paragraph{Acknowledgements} 
%\noindent\textbf{Acknowledgement} 
This work was supported in part by the Chinese Key-Area Research and Development Program of Guangdong Province (2020B0101350001), in part by the Guangdong Basic and Applied Basic Research Foundation (2023A1515011464, 2020B1515020048), in part by the National Natural Science Foundation of China (No.~62102267, No.~61976250), in part by the Shenzhen Science and Technology Program (JCYJ20220818103001002, JCYJ20220530141211024), and the Guangdong Provincial Key Laboratory of Big Data Computing, The Chinese University of Hong Kong, Shenzhen. 

%\clearpage
{\small
\bibliographystyle{ieee_fullname}
\bibliography{egpaper_final}
}

\end{document}